\documentclass[runningheads]{llncs}

\usepackage[lined,algonl,boxed]{algorithm2e}
\usepackage{amssymb}
\usepackage{mismath}

\begin{document}

\title{Combining Sub-Symbolic and Symbolic Methods for Explainability}
%

\institute{Siemens AG, Munich, Germany}

\author{Anna Himmelhuber\inst{1,2} \and Stephan Grimm\inst{1} \and
Sonja Zillner\inst{1} \and
Mitchell Joblin\inst{1} \and Martin Ringsquandl\inst{1} \and
Thomas Runkler\inst{1,2} }
\authorrunning{A. Himmelhuber et al.}
%
\institute{Siemens AG, Munich, Germany \\
\email{\{anna.himmelhuber, stephan.grimm, sonja.zillner, mitchell.joblin, martin.ringsquandl, thomas.runkler\}@siemens.com}
\and
Technical University of Munich, Munich, Germany}
%

%
%

%
\maketitle              
\begin{abstract}

Similarly to other connectionist
models, Graph Neural Networks (GNNs)  lack transparency in their decision-making. A number
of sub-symbolic approaches have been developed to provide insights
into the GNN decision making process. These are first important
steps on the way to explainability, but the generated explanations are
often hard to understand for users that are not AI experts.
To overcome
this problem, we introduce a conceptual approach combining  sub-symbolic and symbolic methods for human-centric explanations, that incorporate domain knowledge and causality. We furthermore introduce the notion of fidelity as a metric for evaluating how close the explanation is to the GNN's internal decision making process. The evaluation with a chemical dataset and ontology shows the explanatory value and reliability of our method. 
\end{abstract}
\section{Introduction}
Many important  real-world  data  sets  come  in  the  form  of graphs or networks, including social networks, knowledge graphs, protein-interaction networks, the World Wide Web and many more. Graph neural networks are connectionist models that capture the  dependence structure induced by links  via message passing between the nodes of graphs. Unlike standard neural networks, GNNs retain a state that can represent  information  from  its  neighborhood  with  arbitrary depth as well as incorporate node feature information \cite{zhou2018graph}. Similarly  to  other  connectionist  models,  GNNs  lack  transparency  in  their  decision-making.  Since the  unprecedented levels of performance of such AI methods lead to increasing use in the daily life of humans, there is an emerging need to understand the decision-making process of such systems \cite{arrieta2020explainable}. While symbolic methods such as inductive logic learning  come with explainability, they perform best when dealing with relatively small and precise data. Sub-symbolic methods such as graph neural networks  are able to handle large datasets, have a higher tolerance to noise in real world data, generally have high computing performance and are easier to scale up \cite{ilkou2020symbolic}.\\
\\
 Through the increasing popularity and need for explainability in AI, a variety of explainable models for neural networks are being developed  \cite{pope2019explainability}. These include surrogate models which are interpretable models that are trained to approximate the predictions of a black box model \cite{ribeiro2016should}.  Other approaches include the identification of the most relevant features \cite{pope2019explainability} \cite{ying2019gnnexplainer}. The explainer methods named above, allow the user to relate properties of the inputs to their output. However, the user is responsible for compiling and comprehending the explanations, relying on their own implicit form of knowledge and reasoning about them.  Since humans are depending on their background knowledge and therefore also their biases about the data and its domain, different explanations about why a model makes a decision may be deduced. Since such sub-symbolic models are often built for AI researchers, it can make them hard to understand for non-experts. We strive to go beyond that by justifying predictions with background or common sense knowledge in a human understandable way \cite{biran2017explanation}. This is of increased importance, as explainable AI and with it the widespread application of AI models are more likely to succeed if the evaluation of these explainer models is focused more on the user's needs \cite{miller2017explainable}.\\
\\
We aim to develop a hybrid method by combining GNNs, sub-symbolic explainer methods and inductive logic learning. This enables human-centric and causal explanations through extracting symbolic explanations from identified decision drivers and enriching them with available background knowledge. These are generated for individual predictions, and are therefore instance-level explanations. With this method, high-accuracy sub-symbolic predictions come with symbolic-level explanations, and provide an effective solution for the performance vs. explainability trade-off. \\
\indent
As far as we know, this is the first work to study integrating a sub-symbolic explainer with symbolic methods for more human-centric instance-level explanations. Our fidelity metric indicates how close an explanation is to the GNN's internal decision making process. Additionally, the employment of justifications in our method provides causality that makes use of the background knowledge.

\section{Background and Problem Definition}\label{background}
\begin{figure}
\vspace{-5mm}
\includegraphics[width=\textwidth]{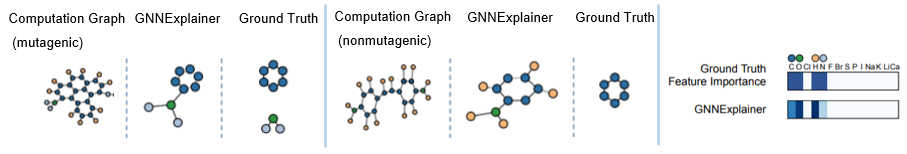}
\caption{GNNExplainer results for Mutag dataset classifications. Figure adapted from Figure 4 and Figure 5 in \cite{ying2019gnnexplainer}.} \label{molecule}
\vspace{-5mm}
\end{figure}

\def\ont{\ensuremath{\mathcal{O}}\xspace}
\def\just{\ensuremath{\mathcal{J}}\xspace}
For incorporating explicit domain knowledge into our explanation method on the side of symbolic representation, we use ontologies expressed in the W3C OWL 2 standard\footnote{\url{https://www.w3.org/TR/owl2-overview/}} \cite{mcguinness2004owl} based on the description logic formalism. In this section we first introduce semantic web ontology, then revisit the notions of entailment, inductive logic learning and justifications, followed by graph neural networks and a sub-symbolic explainer method. Eventually, we define the problem of learning explainer classes by combining a GNN's output with inductive logic learning \footnote{For better readability we will denominate variables represented in ontology form in greek letters and sub-symbolic graph representations in latin letters.}.   \\
\\
\textbf{Semantic Web Ontology}
 The basic constituents for representing knowledge in OWL are individuals, classes and properties. They are used to form axioms, i.e. statements within the target domain, and an ontology \ont is a set of axioms to describe what holds true in this domain. The most relevant axioms for our work are class assertions $\tau(\sigma)$ assigning an individual $\sigma$ to a class $\tau$, property assertions $\rho(\sigma_1,\sigma_2)$ connecting two individuals $\sigma_1,\sigma_2$ by property $\rho$, and subclass axioms $\tau_1 \sqsubseteq \tau_2$ expressing that class $\tau_1$ is a subclass of class $\tau_2$. Classes can be either atomic class names, such as '{\footnotesize \fontfamily{qcr}\selectfont Compound}` or '{\footnotesize \fontfamily{qcr}\selectfont Bond}`, or they can be composed by means of complex class expressions. An example for a complex class expression noted in Manchester syntax is '{\footnotesize \fontfamily{qcr}\selectfont Compound and hasStructure some Nitrogen\_Dioxide}`, which refers to all molecule compounds having some nitrogen dioxide compound in their structure. For details about all types of axioms and the way complex concepts are constructed we refer to \cite{mcguinness2004owl}.\\
 \begin{table}
\vspace{-5mm}
\begin{center}
 \begin{tabular}{p{0.5cm}p{5.7cm}|p{5.5cm}}
 \hline
(1) &  {\footnotesize \fontfamily{qcr}\selectfont Carbon} $\sqsubseteq$ {\footnotesize \fontfamily{qcr}\selectfont Atom} & carbons are atoms\\
(2) & {\footnotesize \fontfamily{qcr}\selectfont Hetero\_aromatic\_5\_ring} $\sqsubseteq$ {\footnotesize \fontfamily{qcr}\selectfont Ring\_Size\_5} $\sqsubseteq$ {\footnotesize \fontfamily{qcr}\selectfont RingStructure} & hetero-aromatic rings of size 5 are rings of size 5, which are ring structures \\
(3) & {\footnotesize \fontfamily{qcr}\selectfont Nitrogen(feature\_100\_5)}  &  feature\_100\_5 is a nitrogen \\
(4) & {\footnotesize \fontfamily{qcr}\selectfont Compound(graph\_100)}  &  graph\_100 is a compound \\
(5) & {\footnotesize \fontfamily{qcr}\selectfont hasAtom(graph\_100, feature\_100\_5)} & graph\_100 has atom feature\_100\_5\\
 \hline
\end{tabular}
\end{center}
\caption{Example excerpt of $\delta^{Mutag}$.}
\vspace{-8mm}
\label{ont}
\end{table}
\\
\textbf{Example 1. (Mutagenesis Ontology)}\\
\textit{As we are combining GNNs and ontologies, graph data has to be available as triples as well as background knowledge. We chose a chemical domain to test our method, as it comes with structured background knowledge. The domain knowledge used in our approach is given by the Mutagenesis ontology $\delta^{Mutag}$~\footnote{ https://github.com/SmartDataAnalytics/DL-Learner/tree/develop/examples\\/mutagenesis}, which is exemplified in Table \ref{ont}}.\\
\textbf{Definition 1. (Entailment)} \\
\textit{Given ontology \ont, if axiom $\alpha$ logically follows from \ont, as can be derived by a standard OWL reasoner, then we call $\alpha$ an \emph{entailment} of \ont and write $\ont \models \alpha$ \footnote{\label{note1}As defined in \cite{horridge2008understanding}}.}\\
\textbf{Definition 2. (Inductive Logic Learning (ILL))}: 
\textit{Given an ontology \ont, a set of positive instances $E^+$ and a set of negative instances $E^-$, a resulting target predicate class expression $\varepsilon$ is constructed such that $\ont \models \varepsilon(\sigma)$ holds for all individuals $\sigma \in E^+$ and does not hold for individuals $\sigma \in E^-$ \footnote{As defined in \cite{lehmann2009dl}}.}\\
\indent
In the context of OWL ontologies, ILL attempts to construct class expressions from an ontology \ont and two sets $E^+,E^-$ of individuals that act as positive and negative examples for being instances of the target class, respectively. 
Concretely, we use DL-Learner \cite{lehmann2009dl} as the key tool to derive OWL class expressions to be used for our explanations.\\
\textbf{Definition 3. (Justification)}:\\
\textit{Given an ontology \ont and an entailment $\alpha$, the \emph{justification} $\just(\ont,\alpha)$ for $\alpha$ in  \ont is a set $\just \subseteq \ont$, such that $\just \models \alpha$ and $\just' \not\models \alpha$ for all proper subsets $\just' \subset \just$ }\\
\textbf{Graph Neural Network (GNN)}\\
For a GNN, the goal is to learn a function of features on a graph \(G=(V, E)\) with edges \(E\) and nodes \(V\). The input is comprised of a feature vector \(x_i\) for every node $i$, summarized in a feature matrix $X \in \mathbb{R}^{n \times d_{in}}$ and a representative description of the link structure in the form of an adjacency matrix $A$. The output of one layer is a node-level latent representation matrix $Z \in \mathbb{R}^{n \times d_{out}}$, where $d_{out}$ is the number of output latent dimensions per node. Therefore, every  layer can be written as a non-linear function:
$ H^{(l+1)} = f(H^{(l)}, A)$,  with  \(H^{(0)} = X\) and  \(H^{(L)} = Z\), $L$ being the number of stacked layers. The vanilla GNN model employed in our framework, uses the propagation rule \cite{kipf2016semi}:
    \[ f(H^{(l)}, A) = \hat{s}(\hat{D}^{-\frac{1}{2}}\hat{A}\hat{D}^{-\frac{1}{2}}H^{(l)}W^{(l)}),\]
    with \(\hat{A} = A + I\), $I$ being the identity matrix. $\hat{D}$ is the diagonal node degree matrix of $\hat{A}$, $W^{(l)}$ is a weight matrix for the $l-th$ neural network layer and $\hat{s}$  is a non-linear activation function.
Taking the latent node representations $Z$ of the last layer we define the logits of node $v_i$ for classification task as $   \hat{y}_i = \text{softmax}(z_i W^{\top}_{c})$,
where $W_c \in \mathbb{R}^{d_{out} \times k}$ projects the node representations into the $k$ dimensional classification space.
\\
\textbf{Example 2. (GNN Classifications for Mutag Dataset)}\\
\textit{
For GNN predictions, the dataset Mutag is utilized, which is from a different source and therefore independent of the Mutagenesis ontology. It contains molecule graphs and is classified through a 3-layer vanilla Graph Convolutional Network with 85\% accuracy \cite{ying2019gnnexplainer}. The molecule graphs $G_i$ = ($A_i$, $X_i$), which are compounds existing out of atoms and bonds, with certain structures such as carbon rings, can be classified as mutagenic (m) or nonmutagenic (n) depending on their mutagenic effect on the Gram-negative bacterium S. typhimurium \cite{debnath1991structure}.}\\
\textbf{Sub-symbolic Explainer Method} \\
The sub-symbolic explainer method  takes a trained GNN and its prediction(s), and it returns an explanation in the form of a small subgraph of the input graph together with a small subset of node features that are most influential for the prediction. For their selection, the mutual information between the GNN prediction and the distribution of possible subgraph structures is maximized through optimizing the conditional entropy. The explainer method output is comprised of edge masks  $M_{E_i} \in  \{0, 1\}^{n \times n}$ $\subset$\ $A_i$ and node feature masks $M_{X_i} \in \{0, 1\}^{n \times d}$ $\subset$\ $X_i$, which is used as input to our framework.  \\
\indent
Since it is the state-of-the-art method, which outperforms alternative baseline approaches by 43.0\% in explanation accuracy \cite{ying2019gnnexplainer},  we chose the GNNExplainer for our framework, but our approach will work with any other explainer subgraph generation method. \\
\textbf{Example 3. (GNNExplainer Output for Mutag Dataset Classifications)}\\
\textit{The GNNExplainer is applied to identify the most influential parts of the respective graph for the classification decision. Figure \ref{molecule} shows the original graph, its edge mask $M_E$ as identified by the GNNExplainer and the ground truth for a mutagenic (left) and nonmutagenic (middle) molecule as well as the identified
node feature mask $M_X$ (right). It can be seen that the identified important graph motifs and node features align with some of ground truth mutagenic properties, as given by \cite{debnath1991structure}. These include ring structures and the node features $C$, $O$, $N$ and $H$. However, the fact that these results represent a carbon ring as well as the chemical group $NO_2$ (Nitrogen dioxide) is left up to the user for interpretation. }
\\
\textbf{Definition 4. (Explainer Class Learning)}\\
\textit{Given ontology $\bigO$ and a set of graph individuals $\{\eta_i\} \in \bigO$~\footnote{Mapping sub-symbolic graph representations ($X_i$, $A_i$) $\mapsto$\ $\bigO$, resulting in individuals $\eta_i$ is specified in Section \ref{learn}} with their respective classifications $\{y_1, y_2,...y_i\}$ provided by a GNN for a certain category, we define  explainer class learning as inductive logic learning such that  $\eta_j|y_j = category\ \in  E^+$ and $\eta_k|y_k \not= category\ \in  E^-$.}  \\
\indent
 $\bigO$ provides the background knowledge for inductive logic learning, and the classification decision by the GNN provides the positive and negative examples in order to learn explainer classes.
We also define a metric called fidelity metric (to be specified
in Section \ref{method}) for quantitative measurement. The higher the fidelity metric, the higher the reliability of the entailed explainer class. 

\section{Combining Sub-Symbolic and Symbolic Methods}\label{method}
\subsection{Explainer Class Learning}\label{learn}
\begin{figure}
\vspace{-1mm}
\includegraphics[width=\textwidth]{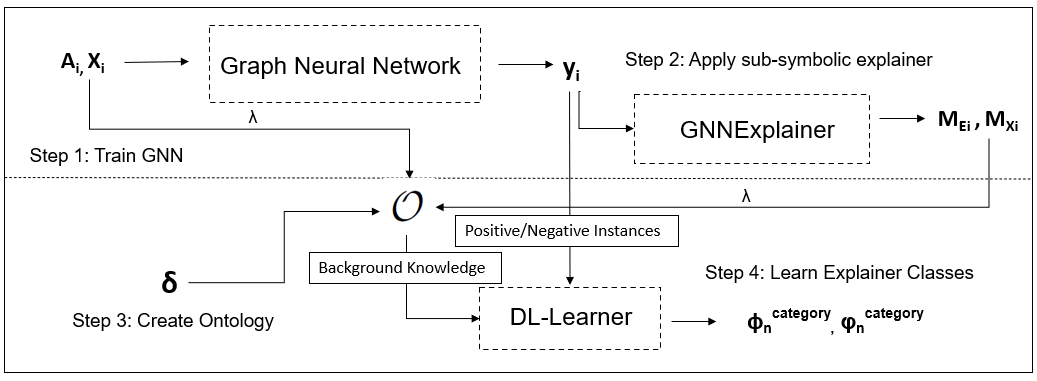}
\caption{Learning Explainer Classes process flow.} \label{swimlanes}
\vspace{-5mm}
\end{figure}
We  are proposing a hybrid method,  within which the coupling of the sub-symbolic explainer method GNNExplainer with the symbolic DL-Learner is used to explain GNN instance-level predictions.  Our approach is shown for a graph classification task, but would equally apply to node classification or link prediction.  The process flow of learning explainer classes can be seen in Figure \ref{swimlanes}.  Firstly, a GNN is trained on and applied to training and testing data and subsequently the sub-symbolic explainer method GNNExplainer is applied to all generated predictions, as can be seen in in Figure \ref{swimlanes} (Step 1 and Step 2). Secondly, to create explainer classes for the GNN decision making process, DL-Learner is applied for a specific predicted category, with positive and negative examples labelled accordingly through $y_i$ (Step 4). The background knowledge used by the DL-Learner to learn explainer classes is comprised of the adjacency matrices $A_i$ and node feature matrices $X_i$, edge masks $M_{E_i}$ and node feature masks $M_{X_i}$ and domain knowledge $\delta$. As the DL-Learner can only process ontologies, the matrices are mapped to an ontology (Step 3) through $\lambda$ as detailed below: \\
\textbf{Extraction and Mapping Step} \\
A set of graphs detailed in their associated matrices $A_i$ and $X_i$~\footnote{Their size is dependent on the number of layers used by the GNN, to keep the consistency in coupling the sub-symbolic with the symbolic method.} are modelled as set of individuals $\{\eta_i\}$. Their edges and node features are extracted from $A_i's$ and $X_i's$ edge and feature lists and modelled as set of individuals $\{\upsilon_j\}$  and $\{\chi_k\}$.
If there are graph-specific structures common in the respective domain,  such as certain motifs, e.g. a ring structure, the set of possible structures $\{structure_z\}$ along with their extraction functions $\{\gamma_z(A_i, X_i)\}$ is defined and mapped through mapping function $S:\{structure_z\}$  $\mapsto$ $\{\gamma_z(A_i, X_i)\}$.\\ 
If $structure_1$ is contained in $(A_i, X_i)$,  extraction function $\gamma_1(A_i, X_i)$ returns all individuals contained in the structure. The found structures are modelled as a set of individuals $\{\psi_g\}$. 
To assign all individuals their type declarations and roles, a set of roles $\{\rho_v\}$ and type declarations $\{\tau_w\}$ as well as further mapping functions based on domain knowledge $\delta$ are needed. Defining these sets and mapping functions has been done as a one-time manual step, with their complexity depending on the domain. \\
$P: \{\eta_i\} \times (\{\upsilon_j\} \cup \{\chi_k\} \cup \{\psi_g\})\mapsto  \{\rho_v\}$, maps a pair of individuals to their role. \\
$T:  (\{\eta_i\} \cup \{\upsilon_j\} \cup \{\chi_k\} \cup \{\psi_g\}\mapsto \{\tau_w\}$  maps individuals to their types. 
All  extracted individuals, roles and type declarations are added as axioms to ontology $\bigO$ through function $AddAxiom(\bigO, axiom)$ as is shown in Algorithm \ref{lambda}.  
Therefore, $\lambda$ is defined as $\lambda(A_i, X_i, T, P, S) \mapsto \bigO$. Equivalently, $\lambda$ is carried out for all corresponding sub-symbolic explainer subgraphs with their associated edge masks $M_{E_i}$ and node feature masks $M_{X_i}$, with the set of explainer graphs modelled as individuals $\eta\_sub_i$.\\
Additionally, mapping function $\mu$ is defined as bijective function, as is shown in Algorithm \ref{lambda}. This function is needed for the fidelity calculation. Function $\mu$ is defined in such a way, that if the input, e.g. ${\sigma_1}$, doesn't map to anything, ${\sigma_1}$ will be returned as output.\\
\textbf{Example 4. (Mapping Mutag Dataset with Mutagenesis Ontology)}\\
\textit{The mapping functions $S^{Mutag} = \{Azanide: \gamma_{Azanide}$ , $Methyl: \gamma_{Methyl}, ...\}$, \\ $R^{Mutag} = \{(\eta_i, \upsilon_j):hasBond, (\eta_i, \chi_k):hasAtom, ...\}$ and \\$T^{Mutag} = \{\eta_i:Compound, \upsilon_j:Bond, \chi_k:Carbon, ...\}$ are defined based on domain terminology $\delta^{Mutag}$. For example, from molecule graph $G_1$ with associated matrices $X_1$ and  $A_1$, the edge individuals edge\_1\_2, edge\_1\_3, etc., are modelled. For extracting structure $Methyl$ ($CH_3$), which is defined as containing one carbon atom bonded to three hydrogen atom, function $\gamma_{Methyl}(A_1, X_1)$ is employed. All accruing axioms are added to the ontology $\bigO^{Mutag}$. Through $\mu$, the set of edges forming the identified structure, e.g. \{ edge\_1\_2,  edge\_1\_3,  edge\_1\_4 \} is mapped to the individual structure\_1\_1\_1.}\\
\\
\begin{algorithm}[H]
    \footnotesize
    \label{lambda}
	\caption{Graph Structure Extraction $\lambda$ } 
	\DontPrintSemicolon
	\KwData{Set of graphs with adjacency matrices $A_i$, feature matrices $X_i$, mapping functions for type declarations $T(\sigma)$, roles $P(\sigma_1, \sigma_2)$ and structures $S(x)$}
\KwResult {$\bigO$, $\mu$}
$\bigO: \{ \}$\;
		\ForEach {graph in range(i)}{{
		       AddAxiom($\bigO$, $T(graph)$ ($\eta_{graph}$))\;}
		\ForEach{edge in\ Edgelist($A_{graph})$}{
	            AddAxiom($\bigO$, $T(edge)$($\upsilon_{edge\_graph}$) )\;
	            AddAxiom($\bigO$, $P(graph, edge)(\eta_{graph}, \upsilon_{edge\_graph})$ )\;}
	   \ForEach{feature in\ Featurelist($X_{graph}$)}{
	         AddAxiom($\bigO$, $T(feature)(\chi_{feature\_graph})$ )\;
	        AddAxiom($\bigO$, $P(graph, feature)(\eta_{graph},\chi_{feature\_graph})$ )\;}
	   \ForEach {structure in\ $\{structure_z\}$} {
	        \If{$S(structure)(A_{graph}, X_{graph})\ not\ \textbf{None}$}{
	       \ForEach {number in range(count($S(structure)(A_{graph}, X_{graph})$))}{
	             AddAxiom($\bigO$, $T(structure)(\psi_{graph\_structure\_number})$ )\;
	       AddAxiom($\bigO$, $P(graph, structure)(\eta_{graph}, \psi_{graph\_structure\_number}))$ \;
	        $\mu: S(structure)(A_{graph}, X_{graph}) \mapsto \psi_{graph\_structure\_number}$
	           }}}}
\end{algorithm}
According to the GNN's classifications positive and negative examples of graphs are distinguished and explainer classes are learned.  The background knowledge is the ontology  $\bigO$ =\ $\delta$\ $\cup$\ $\lambda(A_i, X_i, T, P, S)$\ $\cup$\ $\lambda(M_{E_i}, M_{X_i}, T, P, S)$. We differentiate between two types of explainer classes:   \\
\\
\textbf{Input-Output Explainer Classes} \\
Given Def. 4,  background knowledge $\delta$\ $\cup$\ $\lambda(A_i, X_i, T, P, S)$,\ $\eta_i|y_i = category \in  E^+$ and $\eta_i|y_i \not= category \in  E^-$, a set of Input-Output Explainer Classes $\{\phi_n^{category}\}$ are learned. Input-output explainer classes are candidate explanations, that capture the global behavior of a GNN through investigating what input patterns can lead to a specific class prediction, comparable to the input-output mapping  approach in \cite{sarker2017explaining}. \\
\textbf{Importance Explainer Classes}\\
Given Def.4, background knowledge \ $\delta$\ $\cup$\ $\lambda(M_{E_i}, M_{X_i}, T, P, S)$,  $\eta\_sub_i|y_i = category \in  E^+$ and $\eta\_sub_i|y_i \not= category \in E^-$, a set of Importance Explainer Classes $\{\varphi_m^{category}\}$ are learned. Importance Explainer classes show which edges, nodes, features and motifs are important for the GNN to predict a certain class. These class expressions represent the inner workings of a GNN, by incorporating the output of the sub-symbolic explainer.
\subsection{Explainer Class Application for Instance-Level Explanations}
\begin{figure}

\includegraphics[width=\textwidth]{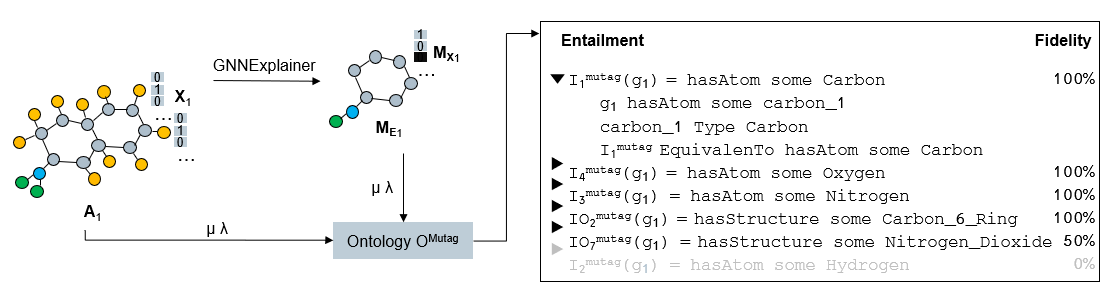}
\caption{Final explanation for molecule graph $G_1$, which has been classified as mutagenic.} \label{mockup}
\vspace{-3mm}
\end{figure}

The pool of possible explainer classes for all categories as learned in Section \ref{learn}, consisting of $\{\phi_n\}$ and $\{\varphi_n\}$, are used in the application step to generate instance-level explanations through explainer class entailment and justification steps. \\
\textbf{Explainer Class Entailment}\\
Given Def. 1., a set of explainer classes  $\{\phi_n^{category}\}$  and $\{\varphi^{category}\}$, ontology $\bigO$ and individual $\eta_j$ classified as category, entailments for  $\eta_j$ are generated.  By doing so, we check if the learned overall decision-making pattern of the GNN applies to a specific instance. For all available explainer classes, entailments for a specific individual $\eta_j$ are generated. It is possible, that several entailments hold, just as it is possible that a classification decision of $G_j$  is based on several different factors. The set of entailments for $\eta_j$ is given by  $C_{Exp}(\eta_j) = \{ \phi\ |\ \bigO \models\  \phi^{category}(\eta_j)\}  \cup \{ \varphi\ |\ \bigO \models\  \varphi^{category}(\eta_j)\}$.  \\
 \textbf{Definition 5. (Entailment Frequency)}\\
 \textit { Given an ontology $\bigO$, explainer class $\phi^{category}_i$ and a set of indivdiuals $\{\eta_i\}$, we define the entailment frequency as the number of entailments for
 $|\{ \eta \in \{\eta_i\}: \bigO \models\ \phi_i^{category}(\eta)\} |$ over the number of instances $|\{\eta_i\}|$.}\\
 \indent
The entailment frequency gives insight over the generality or specificity of explainer classes and representing the average frequency with which a certain explainer class is entailed.
\\
 \textbf{Explainer Class Entailment Justification}\\
 Given $\bigO$ and entailment $\bigO \models\  \phi_i^{category}(\eta_j)$, justification  $\just$($\bigO$, $\phi_i^{category}$($\eta_j$)) is generated.  The number of generated axioms gives some insight about the level of domain knowledge employed. As there can be several justifications for an entailment, we limit them to only one. It is not in the scope of this paper to determine which justification would provide the best explanation, but since a shorter justification tends to be more efficient, the justification with the minimum number of axioms is chosen. \\
 \textbf{Example 5. (Justification for Mutag Explainer Class)}\\
 \textit{Table \ref{just} shows an example justification for the entailment $\bigO^{Mutag}\  \models\  \phi_8^{m}(\eta_1)$, which contributes to a meaningful explanation, as it carries causal information present in expert knowledge about the conclusion.}
 \begin{table}
\vspace{-5mm}
\begin{center}
 \begin{tabular}{p{0.5cm}p{10cm}}
 \hline
(1) & {\footnotesize \fontfamily{qcr}\selectfont $\eta_1$\ hasStructure\ structure\_1\_1\_1} \\
(2) & {\footnotesize \fontfamily{qcr}\selectfont  structure\_1\_1\_1 \ Type\ Hetero\_aromatic\_5\_ring} \\
(3) & {\footnotesize \fontfamily{qcr}\selectfont Hetero\_aromatic\_5\_ring SubClassOf Ring\_size\_5}  \\
(4) & {\footnotesize \fontfamily{qcr}\selectfont $\phi_8^{m}$ EquivalentTo hasStructure some Ring\_size\_5}   \\
 \hline
\end{tabular}
\end{center}
\caption{Example justification $\just$($\bigO^{Mutag}$, $\phi_8^{m}$($\eta_1$)).}
\label{just}
\vspace{-7mm}
\end{table}
\\
\textbf{Fidelity Calculation}\\
Fidelity is defined as the measure of the accuracy of the student model (DL-Learner) with respect to the teacher model (GNN).  High fidelity is therefore fundamental, whenever a student model is to be claimed to offer a good explanation for a teacher model. Without high fidelity, an apparently perfectly good explanation produced by an explainable system is likely not to be an explanation of the underlying sub-symbolic system which it is expected to explain \cite{garcez2020neurosymbolic}. We calculate Fidelity as follows:
 \[ Fidelity(\phi_i, \eta_j) = \frac{|\mu^{-1} (ind(\just(\bigO, \phi_i(\eta_j)))) \cap \eta\_sub_j|}{|\mu^{-1} (ind(\just(\bigO, \phi_i(\eta_j))))| }, \]
where $ind()$ is a function that collects all individuals that are provable instances of a set of axioms. The denominator equals the count of the set of edges or node features that have to be part of $\eta_i$, for the entailment of explainer class $\phi_i$ to hold.
The fidelity metric is defined as the overlap of the sub-symbolic explainer output with the entailed explainer classes, as can be seen in Figure \ref{mockup}, which means that the effectiveness of the sub-symbolic explainer method in representing the GNN decision making is therefore assumed. \\
\textbf{Example 6. (Fidelity for Explainer Class \textit{hasStructure\ some\ Methyl}) }\\
 \textit{As the explainer classes are represented through axioms, e.g.  {\footnotesize \fontfamily{qcr}\selectfont $\phi_2^{n}$ =  hasStructure\ some\ Methyl}, we apply the justification mechanism to arrive at the axioms containing the corresponding individual(s) for the specific example $\eta_1$, such as {\footnotesize \fontfamily{qcr}\selectfont $\eta_1$ hasStructure structure\_1\_2\_1} $\in$ $\just(\phi_2^{n}, \eta_1)$. Since there might be a multiplicity of individuals, function $ind(\just(\bigO, \phi_2(\eta_1))$ is applied, which collects all individuals that are provable instances of the justification. These individuals are then  inversely mapped ($\mu^{-1}$) to their corresponding set of individuals, in this example \{ edge\_1\_2,  edge\_1\_3,  edge\_1\_4 \}. In case there is no corresponding set of individuals, the inverse mapping simply returns the given individual.  For the numerator, we count the overlap of the identified set of individuals with the  individuals in $\eta\_sub_i$, the subgraph identified by the GNNExplainer. }\\
\textbf{Definition 6. (Final Explanation)}\\
\textit{Given the set of entailments, that hold for $\eta_j$, we define the final explanation E($\eta_j$) as the set of the respective justifications  $ E(\eta_j) = {\{\just(\bigO, C(\eta_j))\} \mid C \in C_{Exp(\eta_j)}}$.}
 \\
 \textbf{Example 7. (Molecule Graph $G_1$)}
 \textit{In Figure \ref{mockup}, the final explanation for the classification of molecule graph $G_1$ as mutagenic can be seen, complete with justifications and fidelity score. }
 
\section{Evaluation}
\textbf{Experiment Setting.} We used a subset of 530 molecule graphs as training data to learn explainer classes, and 800 molecule graphs as testing data. The graphs have been classified by a 3-layer vanilla Graph Convolutional Network.  All molecule graphs come with adjacency matrices $A_i^{Mutag}$,  and feature matrices $X_i^{Mutag}$ and their corresponding GNNExplainer importance masks ($M_{E_i}^{Mutag}$ and $X_{E_i}^{Mutag}$), equally split between mutagenic and nonmutagenic classifications. The DL-Learner can create arbitrarily many class expressions, functioning as explainer classes, which are ordered by predictive accuracy (number of correctly classified examples divided by the number of all examples). We are taking a cut-off point of $>50\%$ predictive accuracy, as an explainer class with less than $50\%$ predictive accuracy, wouldn't represent a pattern for mutagenic classification decisions but rather the opposite, and v.v. for nonmutagenic classification decisions~\footnote{All experimental data,  code and results are available from \url{https://github.com/XAI-sub-symbolic/Combining-Sub-Symbolic-Explainer-Methods-with-SWT}.}. \\
\textbf{Explainer Classes}\\
The generated pool of explainer classes provides a total of 14 explainer classes for mutagenic and 12 explainer classes for nonmutagenic classifications. All the comprehensible explanation for mutagenic classification decisions that can be identified and interpreted from the GNNExplainer output (see Section \ref{background}), have been learnt by the DL-Learner. These include   \\
\\
{\footnotesize \fontfamily{qcr}\selectfont $\phi_2^{m}$ =  hasStructure\ some\ Carbon\_6\_ring}, \\
{\footnotesize \fontfamily{qcr}\selectfont $\phi_7^{m}$ = hasStructure some Nitrogen\_dioxide}, \\
{\footnotesize \fontfamily{qcr}\selectfont  $\varphi_1^{m}$ =   hasAtom some Carbon}, \\
{\footnotesize \fontfamily{qcr}\selectfont  $\varphi_2^{m}$ =   hasAtom some Hydrogen},  \\{\footnotesize \fontfamily{qcr}\selectfont  $\varphi_3^{m}$ =   hasAtom some Nitrogen},\\
{\footnotesize \fontfamily{qcr}\selectfont  $\varphi_4^{m}$ =   hasAtom some Oxygen}, \\
\\along with several others, which have not been identified by the GNNExplainer. The explainer class {\footnotesize \fontfamily{qcr}\selectfont  $\phi_6^{m}$ = hasStructure  some Phenanthrene} is a compelling example for the effectiveness of our hybrid approach, as Phenanthrene is a strong indicator for mutagenic potency \cite{debnath1991structure}, but isn't identifiable in the GNNExplainer output. This shows that our hybrid method can identify and verbalize decision-making processes of the GNN, which a comprehensible sub-symbolic explainer system, whose output might not be easily understood and interpreted by a user, is missing. \\
\\
\textbf{Entailment Frequency}
\begin{table}
\vspace{-5mm}
\begin{center}
\begin{tabular}{ |p{2.2cm}|p{1.5cm}|p{2.5cm}|p{2.5cm}|p{2.5cm}|p{2.5cm}|  }
 \hline
  Explainer Class Type &  Number & Avg. Pred. Acc. (SD) & Avg. Entailment Rate (SD) & Avg. Fidelity (SD)\\
 \hline
$\phi_n^{m}$ & 1,...,10 & 0.56 (0.04) & 0.64 (0.3) & 0.88 (0.12)\\
$\phi_n^{n}$ & 1,...,5 & 0.59 (0.03) & 	0.09 (0.04) & 	0.82 (0.12) \\
$\varphi_n^{m}$ & 1,...,4 & 0.77 (0.06) &	0.86 (0.15) & 	0.99 (0.01) \\
$\varphi_n^{n}$ & 1,...,7 & 0.56 (0.01)  & 	0.41 (0.25) & 	0.81 (0.05) \\
 \hline
\end{tabular}
\end{center}
\caption{Input-output and importance explainer classes with avg. pred. accuracy (DL-Learner), entailment rate and fidelity with their respective standard deviations (SD).} 
\label{input}
\vspace{-12mm}
\end{table}\\
\\
The entailment frequency gives us insight over the generality or specificity of explainer classes. As can be seen in Table \ref{input} (Avg. Entailment Rate), there is a wide range of entailment rates.  Some explainer classes, e.g. {\footnotesize \fontfamily{qcr}\selectfont $\phi_4^{m}$ = hasAtom some Carbon} always apply, while others are quite rare, such as {\footnotesize \fontfamily{qcr}\selectfont $\phi_4^{n}$ = hasAtom some Phosphorus}, that comes with only a 4\% entailment rate. As expected, we have an overall lower entailment rate for nonmutagenic explainer classes, as the there are also less distinct factors indicating nonmutagenicity \cite{debnath1991structure}.  Most nonmutagenic classifications come with about 3 entailments, while mutagenic classifications come with more than 5 entailments on average. This is due to a lower generality of the explainer classes, which implies that such an explainer class only applies to specific instances.  This notion is also confirmed by the lower average predictive accuracy of the DL-Learner results for nonmutagenic (57\%) as opposed to mutagenic (63\%) explainer classes, as can be seen in Table \ref{input} (Avg. Pred. Acc). The predictive accuracy of the DL-Learner is defined as the number of correctly classified examples divided by the number of all examples \cite{lehmann2011class}.\\
\\
\textbf{Explanation Fidelity}\\
Fidelity gives the user a measure of reliability of the explanation, with the average fidelity ranging from 64\% for {\footnotesize \fontfamily{qcr}\selectfont $\phi_5^{n}$ = hasStructure some Carbon\_5\_ring} to 100\% for e.g. {\footnotesize \fontfamily{qcr}\selectfont $\varphi_2^{m}$ = hasAtom some Hydrogen}. While an explainer class with an average fidelity of 64\% might still give the user some insight, its explanatory value cannot be considered as reliable as for an explainer class with a higher fidelity. An explainer class, that has a low generality, meaning it is rarely applied to explain a classification, can nonetheless come with a high fidelity such as $\phi_4^{n}$  (100\%). This suggests that also low generality explainer classes can be valuable for specific instances. \\
\\We can observe a positive correlation of 88\% between the average fidelity and predictive accuracy for $\{\varphi_n\}$ and of 50\% between the average fidelity and $\{\varphi_n\}$ $\cup$ $\{\phi_n\}$, signalising the  effectiveness of representing the sub-symbolic decision-making process with the DL-Learner. As the predictive accuracy of the output given by the DL-Learner is the metric on which we base our choice of explainer classes included in the pool, the correlation with the fidelity indicates that this approach leads to reliable explanations. \\
\\Explainability of sub-symbolic methods is desirable not only to justify actions taken based on the predictions made by the system, but also to identify false predictions. Therefore, it is also important to evaluate our method based on its ability to not generate explanation for wrong predictions and therefore validating them. Table \ref{tp} shows the difference in entailments for the correctly classified (true positives TP) and incorrectly classified graphs (false positives FP). We can see, that the average fidelity for entailments is 30 percentage points lower for mutagenic FP than mutagenic TP, and 38 percentage points for nonmutagenic FP.  While this might not be sufficient to clearly identify a wrong classification, it indicates the validity of the fidelity metric, as it is significantly lower for explainer classes applied to incorrect classification.\\
\begin{table}
\vspace{-6mm}
\begin{center}
\begin{tabular}{ |p{3.5cm}|p{2cm}|p{2cm}|p{2cm}|p{2cm}|  }
 \hline
  & $TP^{m}$ & $FP^{m}$ & $TP^{n}$ & $FP^{n}$\\
 \hline
Number of instances & 371 &29 & 374 & 26\\
Average fidelity & 0.96 & 0.66 &0.82 & 0.44\\
 \hline
\end{tabular}
\end{center}
\caption{Average fidelity for true positives and false positives.} 
\label{tp}
\vspace{-7mm}
\end{table}
\\
\textbf{Justification Axioms}\\
Through justifications we provide causality for explanations, based on domain knowledge. The ontology $\delta^{Mutag}$ utilized has little structural depth as can be seen in the example excerpt in Table \ref{ont}. Nonetheless, there is a minimum of 3 axioms for all entailments. For 20\% of explainer classes, 4 axiom justifications and for 8\% of explainer classes, 5 axiom justifications are generated.  This means, that for all explanations generated, the explanations carry some causal information about the conclusion,  supported by expert knowledge.

\subsection{Comparison of our Hybrid Method with DL-Learner Explanations and  Input-Output Explanations}
\textbf{DL-Learner:} 
Classifications along with corresponding explanations can be generated by only using  a symbolic classifier such as the DL-Learner. When comparing this purely symbolic approach with our hybrid method, we find that using only the DL-Learner comes with significantly lower prediction accuracy and also explanatory value. The predictive accuracy of the GNN using the same subset of training data is $78\%$, so considerably above the the DL-Learner result, as shown below. When applying the DL-Learner to carry out classifications, we are restricted to only one classifier. This means, even if we allow more complex class expressions, we only have one explanation for the target predicate mutagenic:\\
       \\
       {\footnotesize \fontfamily{qcr}\selectfont hasStructure\ some\ Nitrogen\_dioxide\ or\ hasThreeOrMoreFusedRings \\ value\ true}  (pred. acc.: 65.76\%) \\
       \\
\textbf{GNN with Input-Output Explanations:}
We want to look at the benefits of integrating a sub-symbolic explainer into our framework, as opposed to explaining GNN predictions with only the input-output matching method as done in e.g. \cite{sarker2017explaining}. We can see, that for some explainer classes such as {\footnotesize \fontfamily{qcr}\selectfont $\phi_3^{m} = \varphi_3^{m}$ = hasAtom some Nitrogen}, we have overlap of the importance explainer classes with the input-output explainer classes. However, the importance explainer classes come with a significantly higher predictive accuracy of 77\% as can be seen in Table \ref{input}, indicating their significance for the classification decision. For the nonmutagenic classifications, explainer class {\footnotesize \fontfamily{qcr}\selectfont $\varphi_2^{n}$ = hasStructure some Carbon\_6\_ring}, which is equivalent with the ground truth as shown in Figure \ref{molecule}, wouldn't have been included in  $\phi_n^{n}$. Here, we can clearly see the added benefit of  generating explainer classes from the GNNExplainer as opposed to only observing the input-output behaviour of a GNN.  The main benefit of including such a sub-symbolic explainer, however, is the provision of the fidelity metric. Without such a metric there is no means to quantify the reliability of the explanation. These results justify the strategy of using a hybrid method. 
\vspace{-5mm}
\\
\subsection{Deeper Integration of GNNs with Domain Knowledge}
\vspace{-2mm}
We carried out an initial integration of sub-symbolic and symbolic methods, by mapping and integrating the GNN input and GNNExplainer output to and with the available domain knowledge. A deeper integration could be reached through integrating available domain knowledge $\delta$ into the GNN before training. As the domain knowledge $\delta$ and the input graphs $G_i$ are from different sources, they are independent. It was therefore not known if their integration could significantly worsen the GNN classification results. Initial results, where we included common molecule structure from domain knowledge $\delta^{Mutag}$ as a simple binary vector into the feature matrices $X_i^{Mutag}$, show that the overall prediction accuracy of the GNN only decreases insignificantly by 2 percentage points, which is a promising first result. It indicates that the domain knowledge, and the explanations generated with it, don't contradict the decision making process of the GNN.

\section{Related Work}
Explainable AI including model-level interpretation and instance-level explanations have been the focus of research for years \cite{biran2017explanation}. In this section we first give an overview for explainable AI for Graph Neural Networks and  then for using symbolic methods to explain sub-symbolic models. \\
\textbf{Sub-Symbolic Explainer Methods}
Current work towards explainable GNNs attempts to convert approaches initially designed for Convolutional Neural Networks (CNNs) into graph domain \cite{pope2019explainability}. The drawback of reusing explanation methods previously applied to CNNs are their inability to incorporate graph-specific data such as the edge structure. Another method, a graph attention model, augments interpretability via an attention mechanism by indicating influential graph structures through learned edge attention weights \cite{2018graph}. It cannot, however, take node feature information into account and is limited to a specific GNN architecture. To overcome these problems, \cite{ying2019gnnexplainer} created the model-agnostic approach GNNExplainer, that finds a subgraph of input data which influence GNNs predictions in the most significant way by maximizing the subgraph's mutual information with the model's prediction.\\
\textbf{Explanations with symbolic methods} A different type of explainability method tries to integrate ML with symbolic methods. The symbolic methods utilized alongside Neural Networks are quite agnostic of the underlying algorithms and mainly harness ontologies and knowledge graphs  \cite{seeliger2019semantic}. One approach is to map network inputs or neurons to classes of an ontology or entities of a knowledge graph. For example, \cite{sarker2017explaining} map scene objects within images to classes of an ontology. Based on the image classification outputted by the Neural Network, the authors run ILP on the ontology to create class expressions that act as model-level explanations.  Furthermore, \cite{selvaraju2018choose} learn a mapping between individual neurons and domain knowledge. This enables the linking of a neuron’s weight to semantically grounded domain knowledge. A ontology-based approach for human-centric explanation of transfer learning is proposed  by \cite{chen2018knowledge}.  While there is some explanatory value to these input-output methods, they fail to give insights into the inner workings of a graph neural network and cannot identify which type of information was influential in making a prediction. This work bridges this gap by combining the advantages of both approaches is among the first to study the coupling of a sub-symbolic explanation method with symbolic methods.
\vspace{-3mm}
\section{Conclusion}
\vspace{-3mm}
In this paper, we addressed the problem of grounding explanations in domain knowledge while keeping them close to the decision making process of a GNN. We showed that combining sub-symbolic with symbolic methods can generate reliable instance-level explanations, that don't rely on the user for correct interpretation. We tested our hybrid framework on the Mutag dataset mapped to the Mutagenesis ontology,  to evaluate its explanatory value, its practicability and the validity of the idea.  We used data from a chemical domain, as it comes with complex domain knowledge that is universally  accepted and can therefore be considered as ground truth when evaluating explanations. Our results show, that there are significant advantages of our hybrid framework over only using the sub-symbolic explainer, where the output is susceptible to biased or faulty interpretations by the user. Equally, there are advantages of our hybrid method over a purely symbolic method such as ILL, as it comes with significantly higher accuracy, while for an input-output method,  the decision-making process of the neural network isn't considered and there are no means to validate the reliability of the explanations. In future, we will evaluate how our hybrid framework compares for different datasets.  Furthermore, we will analyze the effect on explanations when the coupling of available domain knowledge with GNNs is deepened before training. 

%
%
%

\end{document}